\documentclass{article}
\usepackage{ijcai24}

\usepackage{multirow}
\usepackage{pifont}

\usepackage{times}
\usepackage{soul}
\usepackage{url}
\usepackage[hidelinks]{hyperref}
\usepackage[utf8]{inputenc}
\usepackage[small]{caption}
\usepackage{graphicx}
\usepackage{amsmath}
\usepackage{amsthm}
\usepackage{booktabs}
\usepackage{algorithm}
\usepackage{algorithmic}
\usepackage[switch]{lineno}

\title{UIEDP: Underwater Image Enhancement with Diffusion Prior}

\author{
Dazhao Du$^{1}$
\and
Enhan Li$^{1}$\and
Lingyu Si$^{2,3}$\and
Fanjiang Xu$^{2,}$\thanks{Corresponding author.}\and
Jianwei Niu$^{1}$\And
Fuchun Sun$^{2,4}$
\affiliations
$^1$Hangzhou Innovation Institute, Beihang University\\
$^2$Institute of Software, Chinese Academy of Science\\
$^3$University of Chinese Academy of Sciences\\
$^4$Department of Computer Science and Technology, Tsinghua University
\emails
\{dudazhao,lienhan,niujianwei\}@buaa.edu.cn,
\{lingyu,fanjiang\}@iscas.ac.cn,
fcsun@mail.tsinghua.edu.cn
}

\begin{document}

\maketitle

\begin{abstract}
  Underwater image enhancement (UIE) aims to generate clear images from low-quality underwater images. Due to the unavailability of clear reference images, researchers often synthesize them to construct paired datasets for training deep models. However, these synthesized images may sometimes lack quality, adversely affecting training outcomes. To address this issue, we propose UIE with Diffusion Prior (UIEDP), a novel framework treating UIE as a posterior distribution sampling process of clear images conditioned on degraded underwater inputs. Specifically, UIEDP combines a pre-trained diffusion model capturing natural image priors with any existing UIE algorithm, leveraging the latter to guide conditional generation. The diffusion prior mitigates the drawbacks of inferior synthetic images, resulting in higher-quality image generation. Extensive experiments have demonstrated that our UIEDP yields significant improvements across various metrics, especially no-reference image quality assessment. And the generated enhanced images also exhibit a more natural appearance.
\end{abstract}

\section{Introduction}

\begin{figure}[t]
\begin{center}
\includegraphics[width=0.92\columnwidth]{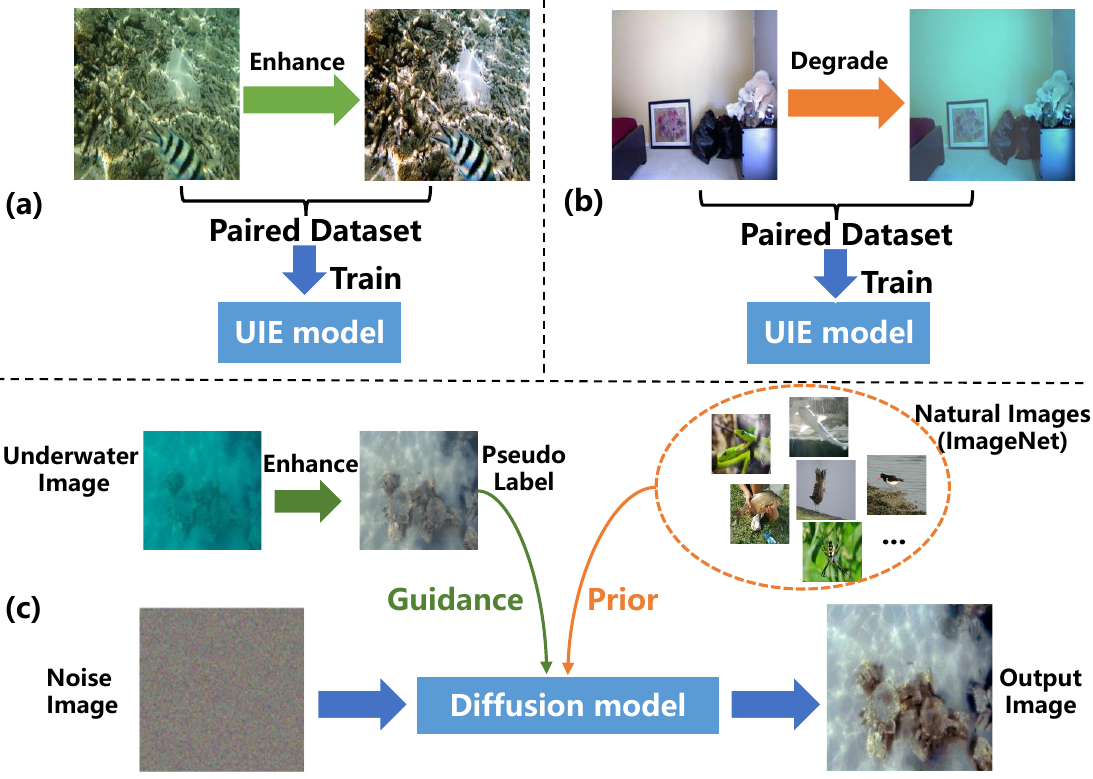}
\caption{(a) Enhance real underwater images to obtain synthesized reference images, and construct paired samples for training UIE models. (b) Degrade natural images to obtain synthesized underwater images, and construct paired samples for training UIE models. (c) A diffusion model trained on ImageNet has natural image priors. To make full use of it, we guide the sampling process of the diffusion model with pseudo-labels generated by any UIE algorithm.}
\label{fig:intro}
\end{center}
\end{figure}

With the rise of marine ecological research and underwater archaeology, processing and understanding underwater images have attracted increasing attention. Due to the harsh underwater imaging conditions, underwater images usually suffer from various visual degradations, such as color casts, low contrast, and blurriness~\cite{fusion,ghani2015enhancement}. To enable advanced visual tasks like segmentation and detection of underwater images, underwater image enhancement (UIE) usually serves as a crucial preprocessing step~\cite{funiegan}.

In recent years, deep learning-based methods have gradually outperformed traditional techniques in UIE. While many deep learning-based UIE models have been proposed~\cite{ucolor,uranker,fivenet}, they heavily rely on paired datasets for training. Due to the unavailability of clean reference images (also known as labels) for real underwater scenes, researchers have employed various approaches to synthesize paired datasets. One approach involves employing multiple UIE algorithms to enhance each underwater image and manually selecting the visually optimal result as the reference image~\cite{waternet,utrans}, as shown in Figure~\ref{fig:intro} (a). However, the upper limit of the reference image quality is limited by the chosen UIE algorithm. The alternative approach utilizes in-air images and corresponding depth maps to generate underwater-style images~\cite{uwgan,uwcnn}, as shown in Figure~\ref{fig:intro} (b). Nevertheless, the synthesized underwater-style images are unreal since the exact degradation model is unknown. To address these issues, some studies have explored domain adaptation to bridge the domain gap between synthetic and real images~\cite{chen2022domain,wang2023domain}. However, such methods often introduce additional discriminators for adversarial training, potentially leading to training instability and mode collapse. Additionally, there are unsupervised~\cite{fu2022unsupervised} and semi-supervised~\cite{huang2023contrastive} approaches that either forgo the use of paired datasets or rely on limited paired samples. However, their performance currently falls short of the state-of-the-art fully supervised methods.

Building upon the success of the diffusion model in various low-level visual tasks~\cite{ddrm,gdp}, we approach UIE as a conditional generation problem. Our objective is to model and sample the posterior probability density of the enhanced image given an underwater image. To achieve this, we propose UIEDP, a novel framework outlined in Figure~\ref{fig:intro} (c). Initially, we employ a diffusion model pre-trained on ImageNet~\cite{imagenet} as our generative model, which can capture the prior probability distribution of natural images. The denoising process of the diffusion model naturally generates realistic images. Subsequently, we utilize the output of any UIE algorithm as a pseudo-label image to guide the conditional generation, and sample from the posterior probability distribution of the enhanced image. If the UIE algorithm of generating pseudo labels is trained on synthetic paired datasets, the introduced diffusion prior can effectively counteract the adverse impact of unreal synthetic images. Furthermore, if the UIE algorithm does not require synthetic paired datasets training, the corresponding UIEDP is considered unsupervised. Experiments demonstrate that benefiting from the diffusion prior, the generated images consistently outperform both pseudo-label images and those produced by other UIE methods across various quality metrics.

In summary, our main contributions are twofold: (1) We design UIEDP, a novel diffusion-based framework for UIE from the perspective of conditional generation. UIEDP can make full use of the generative prior of the pre-trained diffusion model to generate higher-quality enhanced images. Notably, UIEDP is adaptable to both supervised and unsupervised settings, showcasing its versatility. (2) Extensive experiments and ablation studies demonstrate the effectiveness of our framework and illustrate the impact of each component.

\section{Related Work}

\paragraph{Underwater Image Enhancement.} 

Existing UIE methods can be divided into three categories: physical model-based, physical model-free, and data-driven methods. Physical model-based methods estimate the parameters of the underwater image formation model to invert the degradation process. For example, UDCP~\cite{udcp} proposes a variant of DCP~\cite{dcp} that estimates the transmission map through the blue and green color channels, while Galdran \textit{et al.}~\shortcite{galdran2015automatic} proposed a red channel method to recover colors associated to short wavelengths. Sea-thru~\cite{seathru} replaces the atmospheric image formation model with a physically accurate model and restores color based on RGBD images. However, these methods are sensitive to the assumptions made in estimating the model parameters and thus have few applicable scenes and poor robustness. To improve the visual quality, physical model-free methods directly adjust image pixel values by histogram equalization~\cite{histogram}, color balance~\cite{mlle,fusionv2}, and contrast correlation~\cite{ghani2015enhancement}. Fusion~\cite{fusion} is one typical method of these, which produces a more pleasing image by blending color-corrected and contrast-enhanced versions of the original underwater image. However, these methods often suffer from over-enhancement, color distortions, and artifacts when encountered with sophisticated illumination conditions~\cite{Zhuang2022UnderwaterIE}. 

Recently, many works have focused on using deep learning techniques in UIE. However, obtaining reference images for underwater scenes is challenging, leading researchers to employ various methods to synthesize paired datasets. Li \textit{et al.}~\shortcite{uwcnn} synthesized paired data based on underwater scene priors. Additionally, several GAN-based methods~\cite{ugan,funiegan} have been trained adversarially by the generated paired datasets. Li \textit{et al.}~\shortcite{waternet} constructed a paired dataset UIEB by using the best result of several enhancement algorithms to obtain a reference image. Peng \textit{et al.}~\shortcite{utrans} collected a larger dataset LSUI and used it to train a Transformer-based model. These paired datasets have facilitated the development of many UIE models. Some models have incorporated physical priors, such as multi-color space inputs~\cite{ucolor,uiec} and depth maps~\cite{pugan}, to generate visually appealing images. To enhance the efficiency of UIE, Jiang \textit{et al.}~\shortcite{fivenet} proposed a highly efficient and lightweight real-time UIE network with only 9k parameters. In addition to these fully supervised methods, there are also unsupervised~\cite{fu2022unsupervised} and semi-supervised~\cite{huang2023contrastive} methods.

\paragraph{Diffusion Models.} 
Denoising diffusion probabilistic model (DDPM)~\cite{ddpm} is a new generative framework that is utilized to model complex data distributions and generate high-quality images. Based on DDPM, the denoising diffusion implicit model (DDIM)~\cite{ddim} introduces a class of non-Markovian diffusion processes to accelerate the sampling process. In order to achieve conditional generation, Dhariwal \textit{et al.}~\shortcite{classifier} have conditioned the pre-trained diffusion model during sampling with the gradients of a classifier. An alternative way is to explicitly introduce the conditional information as input to the diffusion models~\cite{sr3}. Benefiting from conditional generation, the diffusion models have also been successfully applied to various image restoration tasks~\cite{gdp}, including denoising, deblurring, and inpainting~\cite{DiffPIR,repaint,ddrm}. However, to our best knowledge, only Tang \textit{et al.}~\shortcite{dmuie} have applied a lightweight diffusion model to underwater image enhancement in a supervised manner. In contrast, our approach does not train the diffusion model from scratch but rather makes full use of the natural image priors that are inherent in the pre-trained diffusion model. Furthermore, our method is also applicable to unsupervised scenarios.

\section{Methodology}
\subsection{Preliminary of Diffusion Models}
Denoising diffusion probabilistic model~\cite{sohl2015deep,ddpm} is a type of generative model that converts the isotropic Gaussian distribution into a data distribution. It mainly consists of the diffusion process and the reverse process. The diffusion process is a Markov chain that gradually adds noise to data $x_0$ at $T$ time steps. Each step of the diffusion process can be written as:
\begin{equation}\label{eq:forward}
q(x_{t}|x_{t-1})=\mathcal{N}(x_{t};\sqrt{1-\beta _{t}}x_{t-1},\beta _{t}\mathbf{I}),
\end{equation}
where $\beta_1,...,\beta_{T}$ is a fixed variance schedule. Using the notation $\alpha_t=1-\beta_t$ and $\overline{\alpha}_{t}=\prod_{s=1}^t \alpha _{s}$, we can sample $x_t$ at an arbitrary timestep according to a closed form:
\begin{equation}\label{eq:property}
x_{t}=\sqrt[]{\overline{\alpha} _{t}}x_{0}+\sqrt[]{1-\overline{\alpha} _{t}}\epsilon,
\end{equation}
where $\epsilon$ is Gaussian noise.

To recover the data from the noise, each step of the reverse process can be defined as:
\begin{equation}\label{eq:reverse}
p_{\theta}(x_{t-1}|x_{t})=\mathcal{N}(x_{t-1};\mu_{\theta}(x_{t},t),{\Sigma}_{\theta}\mathbf{I}),
\end{equation}
where $\mu_{\theta}(x_{t},t)$ is the mean, and variance ${\Sigma}_{\theta}$ can be set as learnable parameters~\cite{nichol2021improved} or constants~\cite{ddpm}. By using reparameterization techniques, the mean can be transformed into:
\begin{equation}
\mu_\theta\left(x_t, t\right)=\frac{1}{\sqrt{\alpha_t}}\left(x_t-\frac{\beta_t}{\sqrt{1-\bar{\alpha}_t}} \epsilon_\theta\left(x_t, t\right)\right),
\end{equation}
where $\epsilon_\theta$ is a function approximator intended to predict the noise $\epsilon$ in Equation \ref{eq:property} from $x_t$. By substituting the predicted noise $\epsilon_{\theta}\left(x_{t}, t\right)$ into Equation \ref{eq:property}, we can predict the data $\Tilde{x}_0$ from $x_t$:
\begin{equation}\label{eq:x0}
\tilde{x}_{0} =  \frac{1}{\sqrt{\bar{\alpha}_{t}}}x_{t}-\frac{\sqrt{1-\bar{\alpha}_{t}} }{\sqrt{\bar{\alpha}_{t}}}\epsilon_{\theta}\left(x_{t}, t\right).
\end{equation}

To perform conditional generation, Dhariwal \textit{et al.}~\shortcite{classifier} introduced a classifier $p_{\phi}(y|x_t)$, and each step of the reverse process conditioned on $y$ can be rewritten as:
\begin{equation}\label{eq:guidance}
\begin{split}
    \log p_{\theta}\left(x_{t-1}|x_t, y\right) &=\log \left(p_{\theta}\left(x_{t-1}|x_t\right) p_{\phi}\left(y| x_t\right)\right) + C_1 
    \\  & \approx \log p(z)+C_2,
\end{split}
\end{equation}
where $z \sim \mathcal{N}\left(\mu+\Sigma g, \Sigma \right)$, $\mu$ and $\Sigma$ are the mean and variance of the distribution $p_{\theta}\left(x_{t-1}|x_t\right)$ respectively. $C_1$ and $C_2$ are constants. $g$ is the gradient of the classifier:
\begin{equation}\label{eq:gradient}
    g=\nabla_{x_t} \log p_{\phi}\left(y \mid x_t\right) |_{x_t=\mu}.
\end{equation}
Therefore, the conditional transition operator can be approximated by shifting the mean at each time step by $\Sigma g$.

\subsection{Proposed UIEDP}

\begin{figure}[ht]
\begin{center}
\includegraphics[width=0.85\columnwidth]{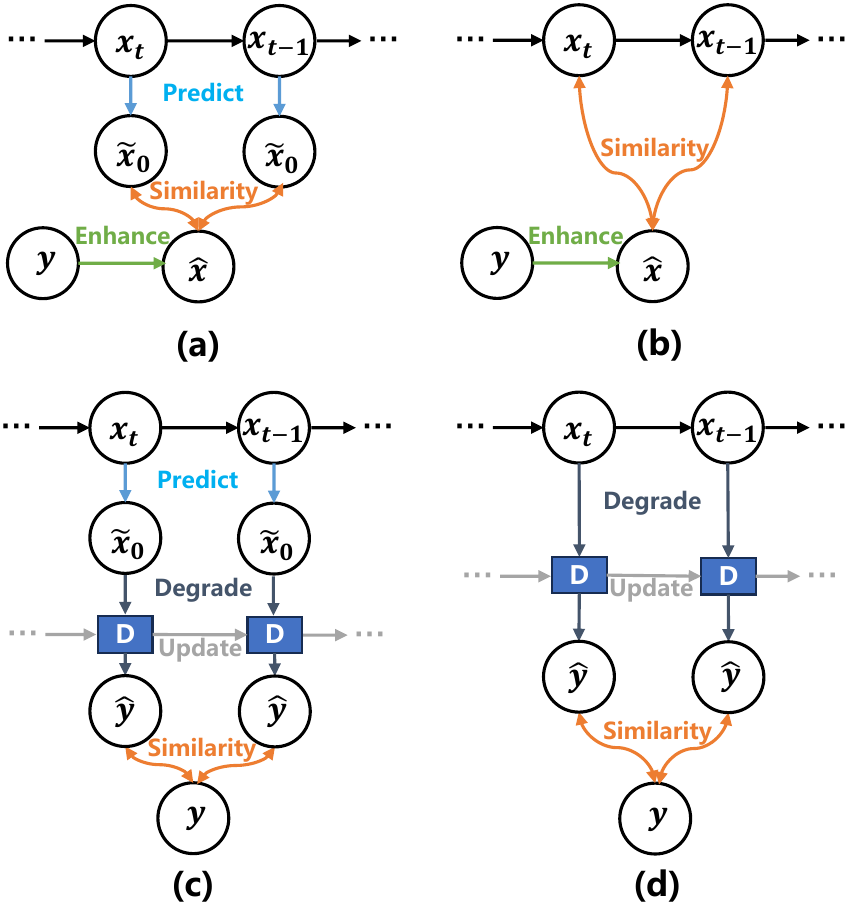}
\caption{Four different guidance methods: guidance on (a) $\tilde{x}_{0}$ and natural image domain, (b) $x_t$ and natural image domain, (c) $\tilde{x}_{0}$ and underwater image domain, (d) $x_t$ and underwater image domain. The blue block D represents the underwater imaging model.}
\label{fig:guide}
\end{center}
\end{figure}

Given an underwater image $y$, the objective of UIE is to generate the enhanced image $x_0$ by searching within the domain of natural images $\mathcal{D}$ to discover a natural image $x_0$ that best matches $y$. Therefore, UIE can be conceptualized as sampling from the posterior probability, $p(x_0|y)$, conditioned on $y$. Due to the stronger generative capabilities demonstrated by diffusion models over other generative models, we utilize a DDPM pre-trained on ImageNet~\cite{imagenet} to establish the prior probability, $p(x_0)$, within the natural image domain. Following Dhariwal \textit{et al.}~\shortcite{classifier} and Equation~\ref{eq:guidance}, we can formulate each step in the reverse process (conditional sampling) as:
\begin{equation}\label{eq:bayes}
    p_{\theta}\left(x_{t-1}|x_t, y\right) = C_3 p_{\theta}\left(x_{t-1}|x_t\right) p\left(y| x_t\right), 
\end{equation}
where $C_3$ is a constant. The former item $p_{\theta}\left(x_{t-1}|x_t\right)$ can be derived from the pre-trained diffusion model, capturing the transition probability from $x_t$ to $x_{t-1}$. $p_{\theta}\left(x_{t-1}|x_t\right)$ is a Gaussian distribution, and we use $\mu$ and $\Sigma$ to denote its mean and variance. Meanwhile, the latter item $p(y|x_t)$ represents the guided item, which can be modeled by a classifier when $y$ corresponds to a specific category. However, in the context of UIE, the role of $p(y|x_t)$ differs. Instead of categorization, we expect $p(y|x_t)$ to quantify the degree of matching between the observed underwater image $y$ and the sampled image $x_t$. Following Avrahami \textit{et al.}~\shortcite{Avrahami_2022_CVPR}, we directly model the negative logarithm of $p(y|x_t)$ using a matching function $\mathcal{L}_{match}(y,x_t)$:
\begin{equation}
    -\log p(y|x_t) = \mathcal{L}_{match}(y,x_t), 
\end{equation}
where the specific implementation of $\mathcal{L}_{match}(y,x_t)$ is left to the following text. Substituting $\mathcal{L}_{match}(y,x_t)$ into Equation~\ref{eq:gradient}, the new gradient of $p(y|x_t)$ is:
\begin{equation}
    g=-\nabla_{x} \mathcal{L}_{match}(y,x_t) |_{x_t=\mu}.
\end{equation}
Therefore, we can shift the mean $\mu$ by $\Sigma g$ during sampling to guide the generation. Next, we explore the design of the matching function $\mathcal{L}_{match}(y,x_t)$, which determines the way guidance is applied.

\paragraph{Guidance on $\mathbf{x_t}$ or $\mathbf{\tilde{x}_{0}}$.}
At each sampling step, in addition to the image $x_t$ sampled from the diffusion model, we also have $\tilde{x}_{0}$ which can be predicted from $x_t$. According to Equation~\ref{eq:x0}, $\tilde{x}_{0}$ is known when $x_t$ is given. Therefore, $p(y|x_t)=p(y|\tilde{x}_{0})$. Similarly, we can use the matching function $\mathcal{L}_{match}(y,\tilde{x}_{0})$ to model the negative logarithm of $p(y|\tilde{x}_{0})$, and the gradient $g$ can be calculated by $g=-\nabla_{\tilde{x}_{0}} \mathcal{L}_{match}(y,\tilde{x}_{0})$. Therefore, we have two different guidance methods: guidance on $x_t$ (Figure~\ref{fig:guide} (b) (d)) and guidance on $\tilde{x}_{0}$ (Figure~\ref{fig:guide} (a) (c)).

Whether the guidance is on $x_t$ ($x=x_t$) or $\tilde{x}_{0}$ ($x=\tilde{x}_{0}$), $x$ belongs to the natural image domain, whereas $y$ belongs to the underwater image domain. Hence, we need to transform $x$ and $y$ into the same domain before measuring the degree of matching $\mathcal{L}_{match}(y,x)$. Similar to Figure~\ref{fig:intro}, there are two implementation approaches. 

\paragraph{Guidance on Natural Image Domain.} We can transform $y$ into the natural image domain to align with $x$. As shown in Figure~\ref{fig:guide} (a) (b), we use any other UIE algorithm to enhance $y$ to generate the pseudo-label image $\hat x$. Then, we calculate the similarity $\mathcal{L}_{sim}(\hat x,x)$ between the generated image and the pseudo-label image as the matching function $\mathcal{L}_{match}(y,x)$. It is worth noting that if the UIE algorithm used to generate the pseudo-label image $\hat x$ does not require supervised training, the corresponding UIEDP is also considered unsupervised.

\paragraph{Guidance on Underwater Image Domain.} We can also transform $x$ into the underwater image domain. Referring to the modified Koschmieder’s light scattering model~\cite{fu2022unsupervised}, the degradation of underwater images can be formulated as:
\begin{equation}\label{eq:degrade}
    \hat y = x * T + (1 - T) * A,
\end{equation}
where $A$ is the global background light and $T$ is the medium transmission map representing the percentage of scene radiance reaching the camera after underwater reflection. For each underwater image, we estimate its $A$ by Gaussian blur~\cite{fu2022unsupervised} and $T$ by GDCP~\cite{gdcp}. However, the estimated transmission map may be inaccurate, resulting in an unrealistic underwater image $\hat y$. Inspired by Fei \textit{et al.}~\shortcite{gdp}, we continually update $T$ by gradient descent at each sampling step. After obtaining $\hat y$, we calculate the similarity $\mathcal{L}_{sim}(y,\hat y)$ between the real underwater image and the synthesized version as the matching function $\mathcal{L}_{match}(y,x)$, as shown in Figure~\ref{fig:guide} (c) (d). As no additional UIE algorithm is required, UIEDP which applies guidance on the underwater image domain is unsupervised.

\begin{figure}[t]
\begin{center}
\includegraphics[width=0.98\columnwidth]{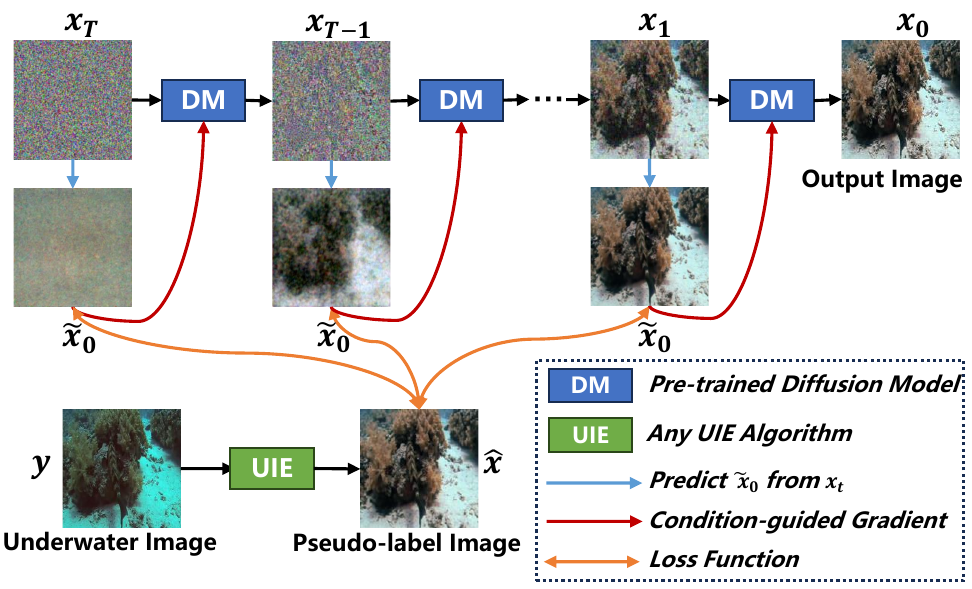}
\caption{The overall framework of UIEDP. Guided by pseudo-labeled images generated from underwater imagery, a pre-trained diffusion model progressively denoises Gaussian noise images, yielding the corresponding enhanced images.}
\label{fig:uiedp}
\end{center}
\end{figure}

By selecting the variables ($x_t$ or $\tilde{x}_{0}$) and domains (natural or underwater) of the guidance, we have four different ways to implement the matching function $\mathcal{L}_{match}(y,x)$, as shown in Figure~\ref{fig:guide}. We have conducted a comprehensive comparison and found that applying guidance on $\tilde{x}_{0}$ and natural image domain (Figure~\ref{fig:guide} (a)) is the best choice, as detailed in Section~\ref{sec:guidance}. Therefore, we implement UIEDP based on this guidance method. 

\begin{algorithm}[t]
    \small
    \caption{UIEDP algorithm}
    \label{alg:uiedp}
    \renewcommand{\algorithmicrequire}{\textbf{Input:}}
    \renewcommand{\algorithmicensure}{\textbf{Output:}}
    \begin{algorithmic}[1]
        \REQUIRE Underwater image $y$, any other UIE algorithm $\textup{Enhance}(\cdot)$, pre-trained diffusion model ($\mu_{\theta}(x_t), \Sigma_{\theta}(x_t), \epsilon_{\theta}(x_t,t)$), loss function $\mathcal L$, and gradient scale $s$
        \ENSURE Enhanced image $x_0$
        \STATE $\hat x \gets \textup{Enhance}(y)$  
        \STATE Sample $x_T\sim \mathcal N (0, \mathbf{I})$
        
        \FOR{$t=T,...,1$}
            \STATE $\mu,\Sigma \gets \mu_{\theta}(x_t), \Sigma_{\theta}(x_t)$ 
            \STATE $\tilde{x}_{0} \gets  \frac{1}{\sqrt{\bar{\alpha}_{t}}}x_{t}-\frac{\sqrt{1-\bar{\alpha}_{t}} }{\sqrt{\bar{\alpha}_{t}}}\epsilon_{\theta}\left(x_{t}, t\right)$ 
            \STATE $g \gets -\nabla_{\tilde{x}_{0}} \mathcal{L}(\hat x, \tilde{x}_{0})$
            \STATE Sample $x_{t-1}\sim \mathcal N (\mu+sg, \Sigma)$
        \ENDFOR
        \RETURN $x_0$
    \end{algorithmic}
\end{algorithm}

As shown in Figure~\ref{fig:uiedp}, we first utilize any other UIE algorithm to transform the underwater image $y$ into the natural image domain, getting the pseudo-label image $\hat x$. Then we predict the mean $\mu$ and $\tilde{x}_{0}$ from the sampled image $x_t$ at each sampling step. To improve the quality of the enhanced image, we add the image quality function of $\tilde{x}_{0}$ to the similarity function $\mathcal{L}_{sim}(\hat x, \tilde{x}_{0})$ between $\hat x$ and $\tilde{x}_{0}$, obtaining the overall loss function $\mathcal L(\hat x, \tilde{x}_{0})$. The gradient of the loss function $g=-\nabla_{\tilde{x}_{0}} \mathcal{L}(\hat x, \tilde{x}_{0})$ is used to guide the conditional generation. Besides, we find that the variance $\Sigma$ negatively influences the generated image~\cite{gdp}. Therefore, we shift the mean $\mu$ by $sg$ instead of $s\Sigma g$, where $s$ is the gradient scale. Algorithm~\ref{alg:uiedp} summarizes the steps of our UIEDP algorithm, and algorithms about other three guidance methods (Figure~\ref{fig:guide} (b) (c) (d)) can be found in the appendix.

\subsection{Loss Function}
The loss function $\mathcal L(\hat x, \tilde{x}_{0})$ is a key factor in guiding conditional generation, mainly composed of two parts: the similarity $\mathcal{L}_{sim}(\hat x, \tilde{x}_{0})$ between $\hat x$ and $\tilde{x}_{0}$, and the quality of $\tilde{x}_{0}$.

\paragraph{Similarity between $\mathbf{\hat x}$ and $\mathbf{\tilde{x}_{0}}$.} In addition to the MAE loss $\mathcal L_{MAE}(\hat x, \tilde{x}_{0})$, the similarity measure also incorporates the multiscale SSIM loss $\mathcal L_{MSSSIM}(\hat x, \tilde{x}_{0})$~\cite{wang2003multiscale} and the perceptual loss $\mathcal L_{Perceptual}(\hat x, \tilde{x}_{0})$~\cite{johnson2016perceptual}. The multiscale SSIM loss mimics the human visual system to measure the multi-scale structural similarity of two images, while the perceptual loss measures the semantic content similarity at the feature level.

\begin{table*}[t]
\centering
\scriptsize
\setlength{\tabcolsep}{1.2mm}{
\begin{tabular}{l|ccccc|ccccc|ccccc}
\toprule
 & \multicolumn{5}{c|}{\textbf{T90}} & \multicolumn{5}{c|}{\textbf{C60}} & \multicolumn{5}{c}{\textbf{U45}} \\
\cmidrule{2-16}   \textbf{Methods} & PSNR$\uparrow$  & SSIM$\uparrow$ & UCIQE$\uparrow$ & UIQM$\uparrow$ & NIQE$\downarrow$ & URanker$\uparrow$ & MUSIQ$\uparrow$ & UCIQE$\uparrow$ & UIQM$\uparrow$ & NIQE$\downarrow$ & URanker$\uparrow$ & MUSIQ$\uparrow$ & UCIQE$\uparrow$ & UIQM$\uparrow$ & NIQE$\downarrow$ \\
\midrule
GDCP    & 13.89 & 0.75 & 0.60 & 2.54 & 4.93 & 0.45 & 46.19 & 0.57 & 2.25 & 6.27 & 0.54 & 47.95 & 0.59 & 2.25 & 4.14\\
Fusion  & 23.14 & 0.89 & \textbf{0.63} & 2.86 & 4.98 & 1.43 & 44.74 & \textbf{0.61} & 2.61 & 5.91 & 1.68 & 47.62 & \textbf{0.64} & 2.97 & 8.02\\
UWCNN   & 19.02 & 0.82 & 0.55 & 2.79 & 4.72 & 0.56 & 42.13 & 0.51 & 2.52 & 5.94 & 0.36 & 46.96 & 0.48 & 3.04 & 4.09\\
UGAN    & 17.42 & 0.76 & 0.58 & \underline{3.10} & 5.81 & 0.68 & 41.12 & 0.55 & 2.86 & 6.90 & 0.89 & 43.96 & 0.57 & 3.10 & 5.79\\
FUnIEGAN  & 16.97 & 0.73 & 0.57 & 3.06 & 4.92 & 0.42 & 41.55 & 0.55 & \textbf{2.87} & 6.06 & 0.80 & 45.68 & 0.56 & 2.92 & 4.45\\
MLLE      & 19.48 & 0.84 & 0.60 & 2.45 & 4.87 & \underline{1.89} & \underline{47.97} & 0.57 & 2.21 & 5.85 & \underline{2.68} & \underline{51.67} & 0.59 & 2.48 & 4.83\\
UShape& 20.24 & 0.81 & 0.59 & 3.06 & \underline{4.67} & 1.11 & 37.67 & 0.56 & 2.73 & 5.60 & 1.57 & 41.10 & 0.57 & 3.19 & 4.20\\
Ucolor  & 20.86 & 0.88 & 0.58 & 3.01 & 4.75 & 0.92 & 45.33 & 0.55 & 2.62 & 6.14 & 1.60 & 47.28 & 0.58 & 3.20 & 4.71\\
UIEC$^2$Net  & \underline{23.26} & \textbf{0.91} & 0.62 & 3.06 & 4.71 & 1.58 & 43.07 & 0.59 & 2.78 & \underline{5.39} & 1.93 & 48.82 & 0.61 & \underline{3.26} & \underline{3.91}\\
NU$^2$Net    & 22.93 & 0.90 & 0.61 & 2.98 & 4.81 & 1.55 & 42.32 & 0.58 & 2.63 & 5.64 & 1.86 & 48.90 & 0.59 & 3.23 & 4.06 \\
FA$^{+}$Net   & 20.98 & 0.88 & 0.59 & 2.92 & 4.83 & 1.26 & 42.41 & 0.57 & 2.52 & 5.70 & 1.59 & 47.50 & 0.58 & 3.21 & 4.02\\
\midrule
UIEDP  & \textbf{23.44} & \textbf{0.91} & \textbf{0.63} & \textbf{3.13} & \textbf{4.54} & \textbf{2.16} & \textbf{51.63} & \underline{0.59} & \textbf{2.87} &  \textbf{5.21} & \textbf{3.58} & \textbf{57.17} & \underline{0.61} & \textbf{3.27} & \textbf{3.79}\\
\bottomrule
\end{tabular}
}
\caption{Comparison with the state-of-the-art results on T90, C60, and U45. \textbf{Bold} and \underline{underlined} denote the best and second-best results in each column, respectively. $\uparrow$ represents the higher is the better as well as $\downarrow$ represents the lower is the better.}
\label{tab:sota}
\end{table*}

\paragraph{Quality of $\mathbf{\tilde{x}_{0}}$.} URanker~\cite{uranker} is a transformer trained on an underwater image quality ranking dataset, capable of producing quality scores for underwater images. MUSIQ~\cite{musiq} is a popular Transformer-based natural image quality assessment (IQA) that has been proven to perform better than other traditional IQA metrics~\cite{huang2023contrastive}. Therefore, we introduce these two differentiable losses: $\mathcal L_{URanker}(\tilde{x}_{0})$ and $\mathcal L_{MUSIQ}(\tilde{x}_{0})$.

The overall loss function $\mathcal L(\hat x, \tilde{x}_{0})$ during sampling is a weighted sum of the above five losses:
\begin{equation}\label{eq:loss} 
\begin{split}
\mathcal L(\hat x, \tilde{x}_{0}) &= \mathcal L_{MAE}(\hat x, \tilde{x}_{0}) - \lambda_1 \mathcal L_{MSSSIM}(\hat x, \tilde{x}_{0}) \\&+ \lambda_2 \mathcal L_{Perceptual}(\hat x, \tilde{x}_{0}) - \lambda_3 \mathcal L_{URanker}(\tilde{x}_{0})\\& - \lambda_4 \mathcal L_{MUSIQ}(\tilde{x}_{0}),
\end{split}
\end{equation}
where $\lambda_1$, $\lambda_2$, $\lambda_3$ and $\lambda_4$ are hyper-parameters to balance the contribution of each loss. 

\section{Experiments}

\subsection{Datasets and Evaluation Metrics}
We empirically evaluate the performance of UIEDP on two public datasets: UIEB~\cite{waternet} and U45~\cite{u45}. The UIEB dataset consists of 890 raw underwater images and corresponding high-quality reference images. Following FA$^{+}$Net~\cite{fivenet}, we use 800 pairs of these images as the training set, and the remaining 90 pairs of images as the test set, named \textbf{T90}. In addition, the UIEB dataset also includes a set of 60 challenge images (\textbf{C60}) that do not have corresponding reference images. The \textbf{U45} dataset contains 45 carefully selected underwater images, serving as an important benchmark for UIE. For all the supervised methods, we train the models on the training set of UIEB and test them on the T90, C60, and U45 datasets. 

In the above three test sets, each underwater image in T90 is paired with a corresponding reference image, while C60 and U45 only contain underwater images. Therefore, we evaluate the performance of UIE using two full-reference image quality evaluation metrics and five non-reference image quality assessment (NR-IQA) metrics. PSNR and SSIM are commonly used full-reference metrics for evaluating generation tasks, measuring the similarity between the enhanced images and the reference images. For non-reference metrics, we have adopted three metrics specifically designed for UIE tasks: UIQM~\cite{UIQM}, UCIQE~\cite{UCIQE}, and URanker~\cite{uranker}, along with two natural image quality assessment metrics: NIQE~\cite{NIQE} and MUSIQ~\cite{musiq}. UIQM considers contrast, sharpness, and colorfulness of the image, while UCIQE considers brightness instead of sharpness. NIQE analyzes the statistical properties of the image to measure the level of noise and distortion. 

\subsection{Implementation Details}

We replicated all the compared methods based on their official codes and hyperparameters. For models requiring supervised training, we trained them for 100 epochs with a batch size of 32, using the ADAM optimizer~\cite{adam} with a learning rate 0.001. Data augmentation techniques such as horizontal flipping and rotation were employed during training. To ensure a fair comparison, all images were resized to 256$\times$256 during both training and testing. Our UIEDP loaded weights of the unconditional diffusion model pre-trained on the ImageNet dataset (at a resolution of 256$\times$256) provided by Dhariwal \textit{et al.}~\shortcite{classifier}. In all experiments, the weights $\lambda_1=1$, $\lambda_2=0.005$, $\lambda_4=0.00001$ in Equation~\ref{eq:loss}. The gradient scale $s$ is set to 12000 on T90 and 4000 on the other two datasets. $\lambda_3$ is set to 0.002 on U45 and 0.001 on the other two datasets.

\begin{figure}[t]
\begin{center}
\includegraphics[width=0.85\columnwidth]{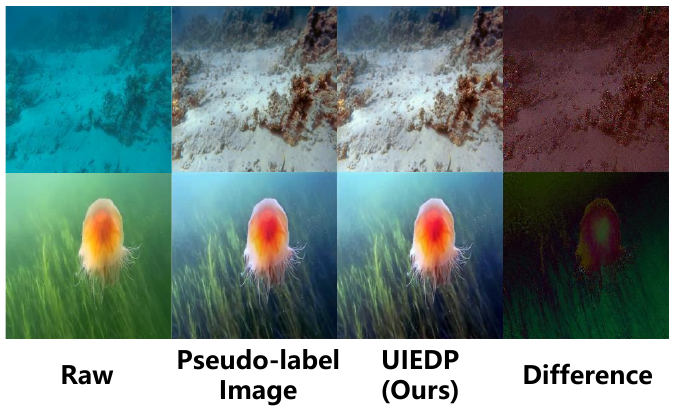}
\caption{Comparison between the pseudo-labeled images and the images generated by UIEDP. The last column denotes their difference images.}
\label{fig:diff}
\end{center}
\end{figure}

\begin{figure*}[t]
\begin{center}
\includegraphics[width=0.9\linewidth]{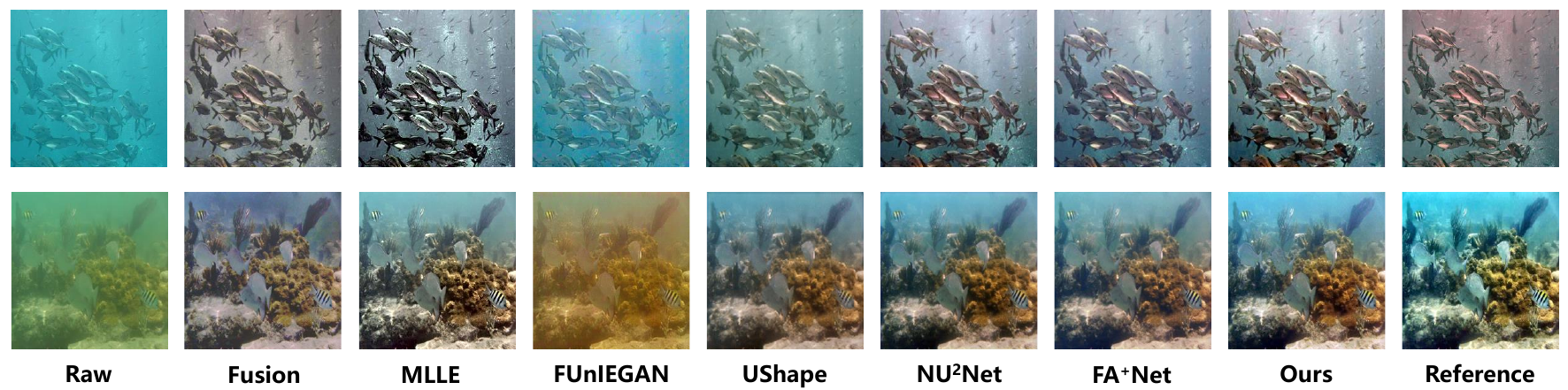}
\caption{Visual comparison of different UIE methods on T90.}
\label{fig:result}
\end{center}
\end{figure*}

\subsection{Comparison with the State-of-the-art}

We compare our method UIEDP with several state-of-the-art approaches, including one physical model-based method (GDCP~\cite{gdcp}), two physical model-free methods (Fusion~\cite{fusion}, MLLE~\cite{mlle}), and eight data-driven methods (UWCNN~\cite{uwcnn}, UGAN~\cite{ugan}, FUnIEGAN~\cite{funiegan}, UShape~\cite{utrans}, UIEC$^2$Net~\cite{uiec}, NU$^2$Net~\cite{uranker}, FA$^+$Net~\cite{fivenet}). For our UIEDP, we employ UIEC$^2$Net trained on the training set to generate pseudo-labeled images.

Results for three datasets are presented in Table~\ref{tab:sota}. Experimental results demonstrate that compared to other methods, we achieve improvements in nearly all metrics across three datasets, particularly in NR-IQA metrics. It's noteworthy that despite employing pseudo-labeled images generated by UIEC$^2$Net for guiding conditional generation, UIEDP outperforms UIEC$^2$Net across all metrics. This phenomenon can be attributed to UIEDP effectively harnessing the natural image priors inherent in the pre-trained diffusion model. Equation~\ref{eq:bayes} indicates that conditional generation is jointly determined by priors and conditional guidance, thereby the introduction of diffusion prior enhances generation quality. Figure~\ref{fig:diff} visually illustrates the difference between output images of UIEDP and pseudo-labeled images. In the first example, the difference lies in the background lighting, with the image generated by UIEDP appearing brighter. In the second example, the difference lies in the jellyfish, where the image generated by UIEDP exhibits enhanced visual appeal.

\begin{table}
    \centering
    \resizebox{0.95\columnwidth}{!}{
    \begin{tabular}{l|ccccc}
    \toprule
     Method & URanker$\uparrow$ & MUSIQ$\uparrow$ & UCIQE$\uparrow$ & UIQM$\uparrow$ & NIQE$\downarrow$ \\
    \midrule
    USUIR & 1.20 & 42.85 & 0.57 & 2.65 & 5.89\\
    Fusion & 1.43 & 44.74 & \textbf{0.61} & 2.61 & 5.91 \\
    UIEDP \small{(Fusion)} & \textbf{2.14} & \textbf{52.69} & \textbf{0.61} & \textbf{2.75} & \textbf{5.80}\\
    \bottomrule
    \end{tabular}
    }
    \caption{Comparison with two unsupervised methods on C60.}
    \label{tab:abla_unsupervised}
\end{table}


Additionally, we present an intuitive comparison with other UIE methods in terms of visual effects. As shown in Figure~\ref{fig:result}, Fusion slightly distorts image colors and blurs image details; FUnIEGAN exhibits severe color distortion; MLLE renders images overly sharp, presenting a black-and-white tone; Ushape generates images with slightly dimmed background light; NU$^2$Net and FA$^+$Net fail to accommodate all underwater scenarios. For example, FA$^+$Net performs well on the first example but poorly on the second. In contrast, the images generated by our UIEDP even appear better than the reference images, correcting the biased red or blue background light. This is attributed to the diffusion prior that is introduced by UIEDP. More visual comparisons are available in the appendix.

Fusion~\cite{fusion} enhances underwater images by adjusting color and contrast without the need for supervised training. Therefore, if we employ Fusion to generate pseudo-labeled images in UIEDP, UIEDP (Fusion) is considered unsupervised. We compare UIEDP with Fusion and another deep learning-based unsupervised method, USUIR~\cite{fu2022unsupervised}, and the experimental results are presented in Table~\ref{tab:abla_unsupervised}. We observe that even USUIR falls short compared to Fusion, indicating the considerable challenge of UIE in an unsupervised setting. Moreover, the performance of UIEDP surpasses both of the two other methods.

\subsection{Ablation Study}

\begin{figure}[t]
\begin{center}
\includegraphics[width=0.98\columnwidth]{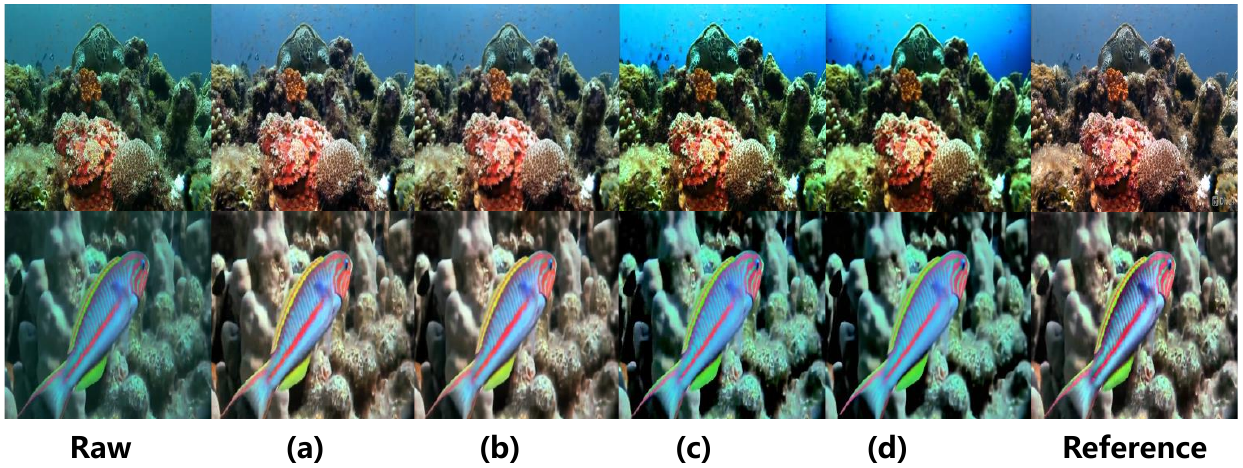}
\caption{Visual comparison of four different guidance methods in Figure~\ref{fig:guide} on T90.}
\label{fig:guide_result}
\end{center}
\end{figure}

\begin{table}
    \centering
    \resizebox{1\columnwidth}{!}{
    \begin{tabular}{l|ccccccc}
    \toprule
    Guidance Method & PSNR$\uparrow$ & SSIM$\uparrow$ & UCIQE$\uparrow$ & UIQM$\uparrow$ & NIQE$\downarrow$ & URanker$\uparrow$ & MUSIQ$\uparrow$ \\
    \midrule
    Fig~\ref{fig:guide} (a) (Ours) & \textbf{23.44} & \textbf{0.91} & 0.62 & \textbf{3.13} & \textbf{4.58} & 2.25 & \textbf{47.63}\\
    Fig~\ref{fig:guide} (b) & 23.06 & 0.89 & 0.62 & 3.10 & 4.94 & \textbf{2.29} & 46.33\\
    Fig~\ref{fig:guide} (c) & 17.50 & 0.76 & \textbf{0.63} & 2.55 & 4.79 & 1.76 & 45.58\\
    Fig~\ref{fig:guide} (d) & 16.93 & 0.73 & \textbf{0.63} & 2.51 & 5.53 & 1.83 & 43.18\\
    \bottomrule
    \end{tabular}}
    \caption{Comparison of different guidance methods in Fig~\ref{fig:guide} on T90.}
    \label{tab:abla_guidance}
\end{table}

\begin{figure*}[t]
\begin{center}
\includegraphics[width=0.85\linewidth]{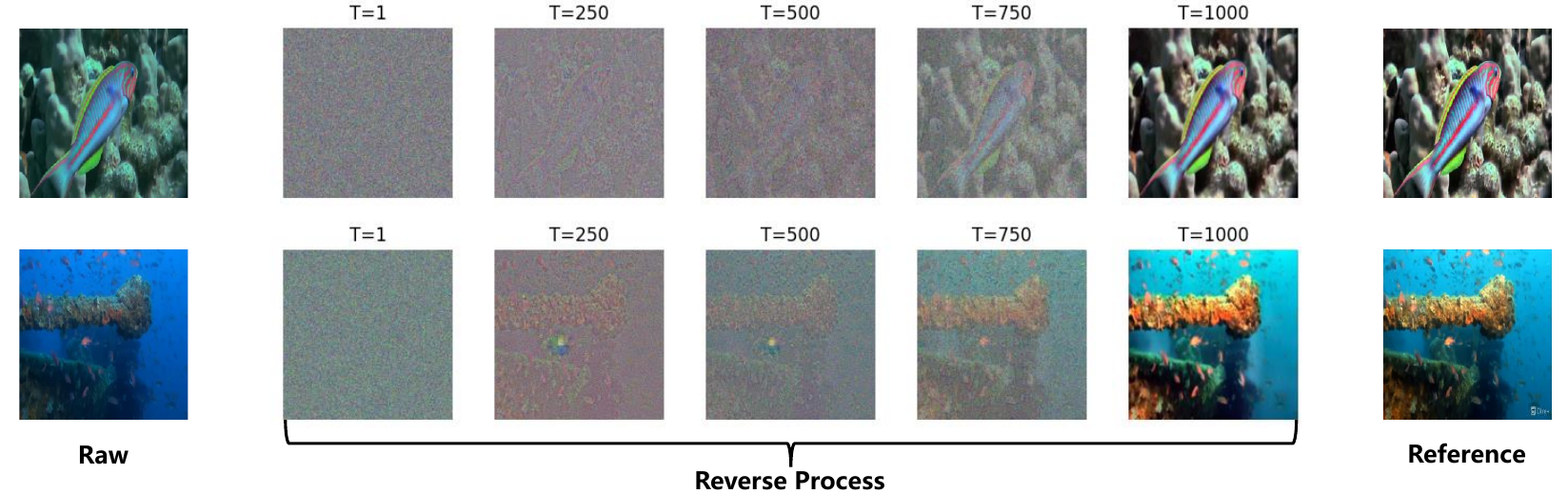}
\caption{Visualization of the inverse diffusion process of UIEDP. The output images at 1000 steps are the final enhanced images.}
\label{fig:diffusion}
\end{center}
\end{figure*}

\paragraph{Ablations about guidance methods.} \label{sec:guidance}

In Figure~\ref{fig:guide}, we propose four different guidance methods. Guidance can act on various variables ($x_t$ or $\tilde{x}_{0}$) and image domains (natural or underwater). To find the optimal guidance method, we conduct both quantitative and qualitative analyses. For fair comparison, we omit quality assessment losses $\mathcal{L}_{MUSIQ}$ and $\mathcal{L}_{URanker}$ in this ablation experiment. As shown in the last two rows of Table~\ref{tab:abla_guidance} and in Figure~\ref{fig:guide_result} (c) (d), guidance in the underwater image domain fails. This is because the degradation model in Equation~\ref{eq:degrade} is a simplified imaging model that cannot precisely model real underwater imaging. And estimated global illumination and transmission maps may also be inaccurate. From Figure~\ref{fig:guide_result} (c) (d), the background light of the enhanced images is not normal. Although Figure~\ref{fig:guide_result} (a) and (b) looks visually similar, results in the first two rows of Table~\ref{tab:abla_guidance} suggest that guidance on $\tilde{x}_{0}$ yields better results than guidance on $x_t$. This is because in the early sampling steps, while $x_t$ still appears as a noisy image, $\tilde{x}_{0}$ already resembles the outline of a normal image, as illustrated in Figure~\ref{fig:uiedp}. From the above analysis, we find that applying guidance on $\tilde{x}_{0}$ and natural image domain (Figure~\ref{fig:guide} (a)) is the optimal guidance method.

\paragraph{Impact of the loss function.} The loss function $\mathcal{L}(\hat x, \tilde{x}_{0})$ plays a critical role in the sampling process, guiding conditional generation through its gradients. In UIEDP, we utilize a weighted sum of five losses as the total loss function $\mathcal{L}(\hat x, \tilde{x}_{0})$. To investigate their contributions, we progressively introduce four additional losses alongside the MAE loss $\mathcal{L}_{MAE}$ for experimental analysis. The comprehensive experimental results are detailed in Table~\ref{tab:abla_loss}. Notably, the inclusion of multiscale SSIM loss $\mathcal{L}_{MSSSIM}$ and perceptual loss $\mathcal{L}_{Perceptual}$ both positively impact the generation process. These losses holistically gauge the similarity between generated images $\tilde{x}_{0}$ and pseudo-labeled images $\hat x$ from various perspectives, leading to enhanced generation quality. Furthermore, MUSIQ $\mathcal{L}_{MUSIQ}$ and URanker $\mathcal{L}_{URanker}$ losses significantly elevate the non-reference quality assessment for the generated images. Additionally, despite conditional generation being guided by pseudo-labeled images, UIEDP can produce higher-quality images compared to the provided pseudo-labeled images. This highlights that the pre-trained diffusion model on ImageNet inherently yields high-quality image outputs, which helps enhance low-quality underwater images.

\begin{table}[t]
    \centering
    \resizebox{1\columnwidth}{!}{
    \begin{tabular}{l|ccccccc}
    \toprule
     Loss Function & PSNR$\uparrow$ & SSIM$\uparrow$ & UCIQE$\uparrow$ & UIQM$\uparrow$ & NIQE$\downarrow$ & URanker$\uparrow$ & MUSIQ$\uparrow$ \\
    \midrule
    Pseudo-label & 23.26 & 0.91 & 0.62 & 3.06 & 4.71 & 2.20 & 47.71 \\
    \midrule
    $\mathcal{L}_{MAE}$ & 23.40 & 0.91 & 0.62 & 3.10 & 4.77 & 2.27 & 47.56\\
    +$\mathcal{L}_{MSSSIM}$ & 23.42 & 0.91 & 0.62 & 3.10 & 4.72 & 2.27 & 47.55\\
    +$\mathcal{L}_{Perceptual}$ & 23.44 & 0.91 & 0.62 & 3.13 & 4.58 & 2.25 & 47.63\\
    +$\mathcal{L}_{URanker}$ & 23.43 & 0.91 & 0.62 & 3.12 & 4.60 & \textbf{2.44} & 47.77\\
    +$\mathcal{L}_{MUSIQ}$ & \textbf{23.44} & \textbf{0.91} & \textbf{0.63} & \textbf{3.13} & \textbf{4.54} & 2.43 & \textbf{50.98}\\
    \bottomrule
    \end{tabular}}
    \caption{Impact of the loss function $\mathcal{L}(\hat x, \tilde{x}_{0})$ on T90. The first row indicates the quality of pseudo-labeled images, and the other rows indicate the quality of enhanced images generated by UIEDP under the guidance of different loss functions.}
    \label{tab:abla_loss}
\end{table}
\begin{table}[t]
    \centering
    \resizebox{0.99\columnwidth}{!}{
    \begin{tabular}{l|c|ccccc}
    \toprule
     Method & Total steps & PSNR$\uparrow$ & SSIM$\uparrow$ & UCIQE$\uparrow$ & UIQM$\uparrow$ & NIQE$\downarrow$ \\
    \midrule

    \multirow{2}{*}{DDIM} 
    & 25 & 21.63 & 0.85 & 0.63 & 2.93 & 5.65\\
    & 50 & 20.08 & 0.79 & \textbf{0.64} & 2.94 & 5.66\\
    \midrule
    \multirow{3}{*}{DDPM} 
    & 50 & 18.50 & 0.66 & 0.61 & 3.06 & 4.95\\
    & 250 & 22.40 & 0.88 & 0.62 & 3.12 & 4.69\\
    & 1000 (Ours) & \textbf{23.44} & \textbf{0.91} & 0.63 & \textbf{3.13} & \textbf{4.54}\\
    \bottomrule
    \end{tabular}}
    \caption{Effect of sampling strategies and steps on T90.}
    \label{tab:abla_ddim}
\end{table}

\paragraph{Effect of sampling strategies and steps.} DDIM~\cite{ddim} improves the sampling process over DDPM, allowing the generation of images of comparable quality with fewer steps. To explore the effect of sampling strategies, we replace the sampling process of UIEDP with DDIM. Experimental results in Table~\ref{tab:abla_ddim} indicate that images sampled via DDIM demonstrate inferior quality compared to DDPM. Additionally, in an effort to expedite generation speed, we attempt to directly reduce sampling steps. For DDIM, appropriately reducing the sampling steps yields better results. However, reducing the sampling steps for DDPM leads to a decrease in the quality of generated images. In Figure~\ref{fig:diffusion}, we visualize the images sampled from the intermediate steps when the total number of steps is 1000. Exploring how to enhance sampling efficiency is left for future research.

\section{Conclusion}

We propose a novel framework UIEDP for underwater image enhancement, which combines the pre-train diffusion model and any UIE algorithm to conduct conditional generation. The pre-trained diffusion model introduces the natural image prior, ensuring the capability to generate high-quality images. Simultaneously, any UIE algorithm provides pseudo-label images to guide conditional sampling, facilitating the generation of enhanced versions corresponding to underwater images. Additionally, UIEDP can work in either a supervised or unsupervised manner, depending on whether the UIE algorithm generating pseudo-labeled images requires paired samples for training. Extensive experiments show that UIEDP not only generates enhanced images of superior quality beyond the pseudo-labeled images but also surpasses other advanced UIE methods.



\bibliographystyle{named}
\bibliography{ijcai24}

\newpage
\appendix
\section{Limitations and Future Work}

Like other diffusion-based methods, the primary drawback of UIEDP is the requirement for many sampling steps, hindering its application in scenarios that demand real-time enhancement. Hence, future research will focus on enhancing the generation efficiency of UIEDP. In the realm of diffusion models, several works have tried to generate high-quality images with fewer sampling steps. Exploring the application of acceleration techniques employed in these studies to enhance the efficiency of UIEDP stands as a promising avenue for exploration.

\section{More Algorithms}

In this section, we elaborate on the implementation of three other guidance methods mentioned in Figure~\ref{fig:guide} of the main paper, along with the integration of DDIM within UIEDP. Algorithms~\ref{alg:uiedp_b},~\ref{alg:uiedp_c},~\ref{alg:uiedp_d} correspond to the three guidance methods depicted in Figure~\ref{fig:guide} (b) (c) (d), respectively. Algorithm~\ref{alg:uiedp_ddim} delineates the steps involved in utilizing DDIM for sampling within UIEDP.

\begin{algorithm}[ht]
    \caption{Guidance on $x_t$ and natural image domain}
    \label{alg:uiedp_b}
    \renewcommand{\algorithmicrequire}{\textbf{Input:}}
    \renewcommand{\algorithmicensure}{\textbf{Output:}}
    \begin{algorithmic}[1]
        \REQUIRE Underwater image $y$, any other UIE algorithm $\textup{Enhance}(\cdot)$, pre-trained diffusion model ($\mu_{\theta}(x_t), \Sigma_{\theta}(x_t)$), loss function $\mathcal L$, and gradient scale $s$
        \ENSURE Enhanced image $x_0$
        \STATE $\hat x \gets \textup{Enhance}(y)$  
        \STATE Sample $x_T\sim \mathcal N (0, \mathbf{I})$
        
        \FOR{$t=T,...,1$}
            \STATE $\mu,\Sigma \gets \mu_{\theta}(x_t), \Sigma_{\theta}(x_t)$ 
            \STATE $g \gets -\nabla_{x_t} \mathcal{L}(\hat x, x_t)$
            \STATE Sample $x_{t-1}\sim \mathcal N (\mu+sg, \Sigma)$
        \ENDFOR
        \RETURN $x_0$
    \end{algorithmic}
\end{algorithm}

\begin{algorithm}[ht]
    \caption{Guidance on $\tilde{x}_{0}$ and underwater image domain}
    \label{alg:uiedp_c}
    \renewcommand{\algorithmicrequire}{\textbf{Input:}}
    \renewcommand{\algorithmicensure}{\textbf{Output:}}
    \begin{algorithmic}[1]
        \REQUIRE Underwater image $y$, global background light $A$, transmission map $T$, learning rate $l$ for optimizing $T$, pre-trained diffusion model ($\mu_{\theta}(x_t), \Sigma_{\theta}(x_t), \epsilon_{\theta}(x_t,t)$), loss function $\mathcal L$, and gradient scale $s$
        \ENSURE Enhanced image $x_0$
        \STATE Sample $x_T\sim \mathcal N (0, \mathbf{I})$
        
        \FOR{$t=T,...,1$}
            \STATE $\mu,\Sigma \gets \mu_{\theta}(x_t), \Sigma_{\theta}(x_t)$ 
            \STATE $\tilde{x}_{0} \gets  \frac{1}{\sqrt{\bar{\alpha}_{t}}}x_{t}-\frac{\sqrt{1-\bar{\alpha}_{t}} }{\sqrt{\bar{\alpha}_{t}}}\epsilon_{\theta}\left(x_{t}, t\right)$ 
            \STATE $\hat y \gets \tilde{x}_{0} * T + (1 - T) * A$
            \STATE $T \gets T - l\nabla_{T} \mathcal{L}(y, \hat y)$
            \STATE $g \gets -\nabla_{\tilde{x}_{0}} \mathcal{L}(y, \hat y)$
            \STATE Sample $x_{t-1}\sim \mathcal N (\mu+sg, \Sigma)$
        \ENDFOR
        \RETURN $x_0$
    \end{algorithmic}
\end{algorithm}

\begin{algorithm}[ht]
    \caption{Guidance on $x_t$ and underwater image domain}
    \label{alg:uiedp_d}
    \renewcommand{\algorithmicrequire}{\textbf{Input:}}
    \renewcommand{\algorithmicensure}{\textbf{Output:}}
    \begin{algorithmic}[1]
        \REQUIRE Underwater image $y$, global background light $A$, transmission map $T$, learning rate $l$ for optimizing $T$, pre-trained diffusion model ($\mu_{\theta}(x_t), \Sigma_{\theta}(x_t)$), loss function $\mathcal L$, and gradient scale $s$
        \ENSURE Enhanced image $x_0$
        \STATE $\hat x \gets \textup{Enhance}(y)$  
        \STATE Sample $x_T\sim \mathcal N (0, \mathbf{I})$
        
        \FOR{$t=T,...,1$}
            \STATE $\mu,\Sigma \gets \mu_{\theta}(x_t), \Sigma_{\theta}(x_t)$ 
            \STATE $\hat y \gets x_t * T + (1 - T) * A$
            \STATE $T \gets T - l\nabla_{T} \mathcal{L}(y, \hat y)$
            \STATE $g \gets -\nabla_{x_t} \mathcal{L}(y, \hat y)$
            \STATE Sample $x_{t-1}\sim \mathcal N (\mu+sg, \Sigma)$
        \ENDFOR
        \RETURN $x_0$
    \end{algorithmic}
\end{algorithm}

\begin{algorithm}[ht]
    \caption{UIEDP algorithm with DDIM sampling strategy}
    \label{alg:uiedp_ddim}
    \renewcommand{\algorithmicrequire}{\textbf{Input:}}
    \renewcommand{\algorithmicensure}{\textbf{Output:}}
    \begin{algorithmic}[1]
        \REQUIRE Underwater image $y$, any other UIE algorithm $\textup{Enhance}(\cdot)$, pre-trained diffusion model $\epsilon_{\theta}(x_t,t)$, loss function $\mathcal L$, and gradient scale $s$
        \ENSURE Enhanced image $x_0$
        \STATE $\hat x \gets \textup{Enhance}(y)$  
        \STATE Sample $x_T\sim \mathcal N (0, \mathbf{I})$
        
        \FOR{$t=T,...,1$}
            \STATE $\tilde{x}_{0} \gets  \frac{1}{\sqrt{\bar{\alpha}_{t}}}x_{t}-\frac{\sqrt{1-\bar{\alpha}_{t}} }{\sqrt{\bar{\alpha}_{t}}}\epsilon_{\theta}\left(x_{t}, t\right)$ 
            \STATE $g \gets -\nabla_{\tilde{x}_{0}} \mathcal{L}(\hat x, \tilde{x}_{0})$
            \STATE $\hat \epsilon \gets \epsilon_{\theta}(x_t,t) - \sqrt{1-\bar{\alpha}_{t}}sg$ 
            \STATE $x_{t-1} \gets \sqrt{\bar{\alpha}_{t-1}}(\frac{x_t-\sqrt{1-\bar{\alpha}_{t}}\hat{\epsilon}}{\sqrt{\bar{\alpha}_t}}) + \sqrt{1-\bar{\alpha}_{t-1}}\hat \epsilon$ 
        \ENDFOR
        \RETURN $x_0$
    \end{algorithmic}
\end{algorithm}

\section{More ablations}
In this section, we conducted more ablations to further validate the effectiveness of UIEDP.

\paragraph{Ablations about other UIE algorithms.} In the main paper, we utilize UIEC$^2$Net and Fusion as the algorithms responsible for generating pseudo-labeled images in UIEDP. We observe consistent improvements in image quality generated by UIEDP compared to the pseudo-labeled images. Consequently, we replace the algorithms generating pseudo-labeled images with MLLE and FA$^+$Net to ascertain if this improvement persisted. The experimental results are presented in Table~\ref{tab:abla_uiedp}. Our findings indicate that irrespective of the UIE algorithm integrated into UIEDP, it consistently generates superior enhanced images, which further emphasizes the effectiveness of the diffusion prior.

\paragraph{Ablations about the variance $\Sigma$.} During the conditional sampling process, we omitted one coefficient of gradient, i.e., the distribution variance $\Sigma$ predicted by the diffusion model. According to the experimental results in Table~\ref{tab:abla_variance}, we find that the variance $\Sigma$ negatively influences the generated images. Hence, we shift the mean by $sg$ instead of $s\Sigma g$.

\begin{figure*}[t]
\begin{center}
\includegraphics[width=0.9\linewidth]{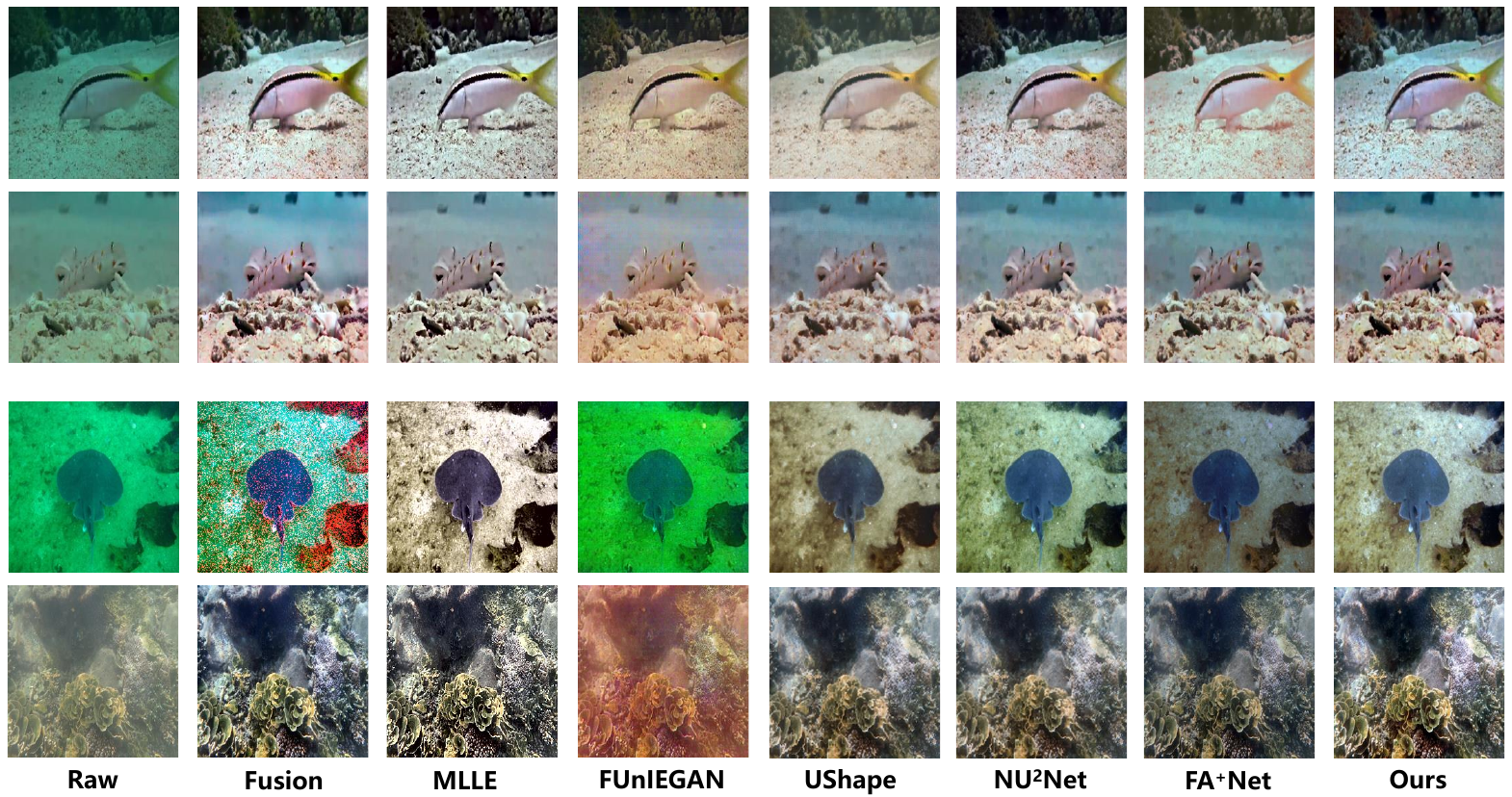}
\caption{Visual comparison of different UIE methods on C60 and U45.}
\label{fig:result1}
\end{center}
\end{figure*}

\begin{table}[t]
    \centering
    \resizebox{1\columnwidth}{!}{
    \begin{tabular}{l|ccccc}
    \toprule
     & PSNR$\uparrow$ & SSIM$\uparrow$ & UCIQE$\uparrow$ & UIQM$\uparrow$ & NIQE$\downarrow$ \\
    \midrule
    MLLE  & 19.48 & 0.84 & \textbf{0.60} & 2.45 & 4.87\\
    UIEDP \small{(MLLE)} & \textbf{19.91} & \textbf{0.85} & \textbf{0.60} & \textbf{2.53} & \textbf{4.65}\\
    \midrule
    FA$^+$Net  & 20.98 & \textbf{0.88} & 0.59 & 2.92 & 4.83\\
    UIEDP \small{(FA$^+$Net)} & \textbf{21.05} & \textbf{0.88} & \textbf{0.60} & \textbf{3.05} & \textbf{4.64}\\
    \bottomrule
    \end{tabular}}
    \caption{Ablations about other UIE algorithms on T90.}
    \label{tab:abla_uiedp}
\end{table}

\begin{table}[t]
    \centering
    \resizebox{1\columnwidth}{!}{
    \begin{tabular}{l|ccccc}
    \toprule
     & PSNR$\uparrow$ & SSIM$\uparrow$ & UCIQE$\uparrow$ & UIQM$\uparrow$ & NIQE$\downarrow$ \\
    \midrule
    with $\Sigma$  & 22.98 & 0.90 & 0.62& 3.03 & 4.60\\
    without $\Sigma$ & \textbf{23.44} & \textbf{0.91} & \textbf{0.63} & \textbf{3.13} & \textbf{4.54}\\
    \bottomrule
    \end{tabular}}
    \caption{Ablations about the variance $\Sigma$ on T90.}
    \label{tab:abla_variance}
\end{table}

\section{More Visual Results}

As shown in Figure~\ref{fig:result1}, we present an intuitive comparison with other UIE methods in terms of visual effects. The first two rows are samples from C60, and the last two rows are samples from U45.

\end{document}